%% file: ms.tex
\documentclass[sigconf,authorversion,screen]{acmart}

\usepackage{array}
\usepackage{amsmath}
\usepackage{multirow}

\usepackage[ruled]{algorithm2e}

\SetAlFnt{\small}
\SetAlCapFnt{\small}
\SetAlCapNameFnt{\small}
\SetAlCapHSkip{0pt}

\copyrightyear{2021}
\acmYear{2021}
\setcopyright{acmlicensed}
\acmConference[MM '21]{Proceedings of the 29th ACM International Conference on Multimedia}{October 20--24, 2021}{Virtual Event, China}
\acmBooktitle{Proceedings of the 29th ACM International Conference on Multimedia (MM '21), October 20--24, 2021, Virtual Event, China}
\acmPrice{15.00}
\acmDOI{10.1145/3474085.3475497}
\acmISBN{978-1-4503-8651-7/21/10}

\graphicspath{{figs/}}

\input{macro.tex}

\begin{document}
\fancyhead{}

%%%%%%%%% TITLE
\title{Constrained Graphic Layout Generation via Latent Optimization}
% LayoutGAN++
% Generative Adversarial Layout Transformers with On-demand Design Constraints

%%%%%%%%% AUTHOR
\author{Kotaro Kikuchi}
\affiliation{
  \institution{Waseda University}
  \city{Shinjuku-ku}
  \state{Tokyo}
  \country{Japan}
}
\email{kiku-koh@ruri.waseda.jp}

\author{Edgar Simo-Serra}
\affiliation{
  \institution{Waseda University}
  \city{Shinjuku-ku}
  \state{Tokyo}
  \country{Japan}
}
\email{ess@waseda.jp}

\author{Mayu Otani}
\affiliation{
  \institution{CyberAgent\del{, Inc.}}
  \city{Shibuya-ku}
  \state{Tokyo}
  \country{Japan}
}
\email{otani_mayu@cyberagent.co.jp}

\author{Kota Yamaguchi}
\affiliation{
  \institution{CyberAgent\del{, Inc.}}
  \city{Shibuya-ku}
  \state{Tokyo}
  \country{Japan}
}
\email{yamaguchi_kota@cyberagent.co.jp}

\renewcommand{\shortauthors}{Kikuchi, Simo-Serra, Otani, and Yamaguchi}

%%%%%%%%% ABSTRACT
\begin{abstract}
% Visual communication is essential to our daily lives, and is established through the graphic layout.
% Creating a layout that satisfies a given set of constraints is a frequent part of the design process.
% The challenge is that these constraints are not invariant and must be flexibly revised on demand.
% Previous works in layout generation trained generative adversarial networks (GANs) with additional losses to satisfy constraints, which is impractical as it requires re-training when the constraints change.
% In this work, we formulate the constrained layout generation task as constrained latent code optimization using a pre-trained GAN.
% The proposed framework is capable of generating layouts that satisfy various constraints using the identical GAN.
% Our proposed optimization framework and a well-tuned Transformer-based model, named LayoutGAN++,
% achieved fascinating results in publicly available benchmarks for both constrained and unconstrained layout generation tasks.
It is common in graphic design humans visually arrange various elements
according to their design intent and semantics. For example, a title text
almost always appears on top of other elements in a document. In this work, we
generate graphic layouts that can flexibly incorporate such design semantics,
either specified implicitly or explicitly by a user. We optimize using the
latent space of an off-the-shelf layout generation model, allowing our approach
to be complementary to and used with existing layout generation models.  Our
approach builds on a generative layout model based on a Transformer
architecture, and formulates the layout generation as a constrained
optimization problem where design constraints are used for element alignment,
overlap avoidance, or any other user-specified relationship.  We show in the experiments
that our approach is capable of generating realistic layouts in both
constrained and unconstrained generation tasks with a single model.
\add{The code is available at \url{https://github.com/ktrk115/const_layout}.}
\end{abstract}




\begin{teaserfigure}
  \vspace*{-3mm}
  \begin{center}
    \includegraphics[width=\textwidth]{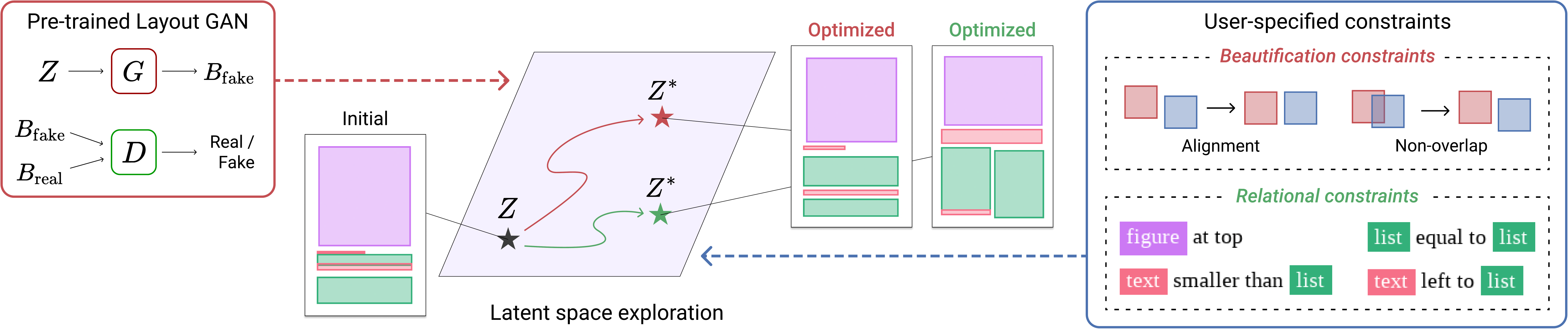}
  \end{center}
  \vspace*{-3mm}
  \caption{Overview of our \del{proposed} Constrained Layout Generation via Latent Optimization (CLG-LO) framework.
  Given a pre-trained Generative Adversarial Network (GAN) for layout generation and user-specified constraints, CLG-LO explores the latent code to find a layout that satisfies the constraints.
  CLG-LO can reuse the same GAN for varying constraints without re-training.
  % otani: following sentence can be omitted
  %We showcase layout generation with beautification constraints and relational constraints as in the examples. 
  }
  \label{fig:teaser}
\end{teaserfigure}

\maketitle

\input{figtab.tex}

%%%%%%%%% BODY TEXT
\input{intro.tex}
\input{related.tex}
\input{approach.tex}
\input{experiment.tex}
\input{conclusion.tex}

%\bibliographystyle{ACM-Reference-Format}
%\bibliography{reference}
\input{ms.bbl}

\end{document}

%% file: macro.tex
\newcolumntype{C}[1]{>{\centering\arraybackslash}m{#1}}
\newcolumntype{L}[1]{>{\arraybackslash}m{#1}}

\newcommand{\ie}{\textit{i}.\textit{e}.}
\newcommand{\eg}{\textit{e}.\textit{g}.}

\newcommand{\std}[1]{\scriptsize $\pm$#1}

\newcommand{\ch}{$\checkmark$}
\newcounter{index}

\newcommand{\add}[1]{#1}
\newcommand{\del}[1]{}

%% file: figtab.tex
%%%% Dataset stats.
\newcommand{\TabDataset}{
\setlength\tabcolsep{3.5pt}
\begin{table}
\caption{Statistics of the datasets used in our experiments and the splits using for evaluation.}
\label{tab:dataset}
\vspace*{-2mm}

\begin{center} \begin{tabular}{l C{10mm}C{14mm} rrr} \toprule
Dataset & \#~label types & Max. \#~elements & \# train. & \# val. & \# test. \\
\midrule
Rico \cite{Deka2017,Liu2018} & 13 & 9 & 17,515 & 1,030 & 2,061 \\
PubLayNet \cite{Zhong2019} & 5 & 9 & 160,549 & 8,450 & 4,226 \\
Magazine \cite{Zheng2019} & 5 & 33 & 3,331 & 196 & 392 \\
\bottomrule \end{tabular} \end{center} \end{table}
\setlength\tabcolsep{6pt}
}

%%%% FID comparison
\newcommand{\fidfigs}[1]{
\frame{\includegraphics[width=9mm]{fid/#1_0.png}} &
\frame{\includegraphics[width=9mm]{fid/#1_1.png}} &
\frame{\includegraphics[width=9mm]{fid/#1_2.png}} &
\frame{\includegraphics[width=9mm]{fid/#1_3.png}}
}
\newcommand{\TabFID}{
\setlength\tabcolsep{0.5pt}
\begin{table}
\caption{Comparison of FID scores computed using feature extractors trained
with various objectives. In particular we compare feature extractors trained
with classification loss (Class\del{.}), \del{R}\add{r}econstruction loss (Recon\del{.}), and a
combination of both (Class\del{. }+\del{ }Recon\del{.}). We compute the FID score between real
layouts and variants that have added noise, have been vertically flipped, and
nearest neighbo\del{u}rs from the validation set.}
\label{tab:compare_fid}
\vspace*{-2mm}

\small{ \begin{center} \begin{tabular}{C{12mm} C{9.5mm}C{9.5mm}C{9.5mm}C{9.5mm} C{10mm}C{10mm}C{10mm}}
\toprule
\multicolumn{5}{c}{Layout variants} & Class\del{.} & Recon\del{.} & Class\del{. +}\add{+} Recon\del{.} \\
\midrule
Real & \fidfigs{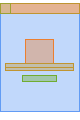} & - & - & - \\
Added noise & \fidfigs{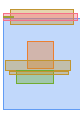} & 186.64 & 37.99 & 127.57 \\
Vertically flipped \hspace{1mm} & \fidfigs{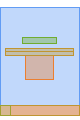} & 3.37 & 97.91 & 100.34 \\
Nearest neighbour & \fidfigs{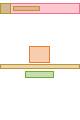}  & 0.29 & 12.52 & 11.80 \\
\bottomrule \end{tabular} \end{center} } \end{table}
\setlength\tabcolsep{6pt}
}

%%%% Unconstrained layout generation

%%% ULG table
\newcommand{\ULGTable}{
\setlength{\tabcolsep}{2.6pt}
\begin{table*} \begin{center} 

\caption{Quantitative comparison of unconstrained layout generation. The values
of Alignment and Overlap are multiplied by $100\times$ for visibility.
Comparisons are provided on three different datasets (Rico, PubLaynet, and
Magazine). For reference, the FID and Max. IoU computed between the validation
and test data, and the Alignment and Overlap computed with the test data are shown as \emph{real data}.}
\label{tab:ulg_result}
\vspace*{-2mm}

\footnotesize{ \begin{tabular}{l cccc cccc cccc} \toprule
\multicolumn{1}{r}{Dataset} & \multicolumn{4}{c}{Rico} & \multicolumn{4}{c}{PubLayNet} & \multicolumn{4}{c}{Magazine} \\
\cmidrule(lr){2-5} \cmidrule(lr){6-9} \cmidrule(lr){10-13}
Model
& FID $\downarrow$ & Max. IoU $\uparrow$ & Alignment $\downarrow$ & Overlap $\downarrow$
& FID $\downarrow$ & Max. IoU $\uparrow$ & Alignment $\downarrow$ & Overlap $\downarrow$
& FID $\downarrow$ & Max. IoU $\uparrow$ & Alignment $\downarrow$ & Overlap $\downarrow$ \\
\midrule

LayoutGAN-W \cite{Li2019}
& 162.75\std{0.28} & 0.30\std{0.00} & 0.71\std{0.00} & 174.11\std{0.22}
& 195.38\std{0.46} & 0.21\std{0.00} & 1.21\std{0.01} & 138.77\std{0.21}
& 159.20\std{0.87} & 0.12\std{0.00} & {\bf 0.74}\std{0.02} & 188.77\std{0.93}
\\

LayoutGAN-R \cite{Li2019}
& 52.01\std{0.62} & 0.24\std{0.00} & 1.13\std{0.04} & 69.37\std{0.66}
& 100.24\std{0.61} & 0.24\std{0.00} & 0.82\std{0.01} & 45.64\std{0.32}
& 100.66\std{0.35} & 0.16\std{0.00} & 1.90\std{0.02} & 111.85\std{1.44}
\\

NDN-none \cite{Lee2020}
& {\bf 13.76}\std{0.28} & 0.35\std{0.00} & {\bf 0.56}\std{0.03} & {\bf 54.75}\std{0.29}
& 35.67\std{0.35} & 0.31\std{0.00} & 0.35\std{0.01} & {\bf 16.50}\std{0.29}
& 23.27\std{0.90} & 0.22\std{0.00} & 1.05\std{0.03} & {\bf 30.31}\std{0.77}
\\

LayoutGAN++
& 14.43\std{0.13} & {\bf 0.36}\std{0.00} & 0.60\std{0.12} & 59.85\std{0.59}
& {\bf 20.48}\std{0.29} & {\bf 0.36}\std{0.00} & {\bf 0.19}\std{0.00} & 22.80\std{0.32}
& {\bf 13.35}\std{0.41} & {\bf 0.26}\std{0.00} & 0.80\std{0.02} & 32.40\std{0.89}
\\

\midrule

Real data
& 4.47 & 0.65 & 0.26 & 50.58
& 9.54 & 0.53 & 0.04 & 0.22
& 12.13 & 0.35 & 0.43 & 25.64
\\

\bottomrule \end{tabular} }
\end{center} \end{table*}
\setlength\tabcolsep{6pt}
}

%%% ULG ablation table
\newcommand{\ULGAblTable}{
\setlength\tabcolsep{3pt}

\begin{table*} \begin{center}
\caption{Ablation study on Rico. The values of Alignment and Overlap are
multiplied by $100\times$ for visibility.}
\label{tab:ablation_study_rico}

\begin{tabular}{l cc cc c cccc} \toprule
& \multicolumn{2}{c}{Self-attention} & \multicolumn{2}{c}{Discriminator} & & \multicolumn{4}{c}{Rico} \\
\cmidrule(lr){2-3} \cmidrule(lr){4-5} \cmidrule(lr){7-10}
Model & Stacked & Transformer & Max pool & Special token & Recon.
& FID $\downarrow$ & Max. IoU $\uparrow$ & Alignment $\downarrow$ & Overlap $\downarrow$ \\
\midrule

% ablation2_seed_0_SRLN_MP_WO
& \ch & & \ch & &
& 197.3\std{0.6} & 0.179\std{0.001} & 0.776\std{0.007} & 278.340\std{0.455} \\

% ablation2_seed_0_SRLN_MP_W
& \ch & & \ch & & \ch
& 32.3\std{0.6} & 0.329\std{0.002} & 0.916\std{0.014} & 78.285\std{0.476} \\

% ablation2_seed_0_SRLN_ST_WO
& \ch & & & \ch &
& 115.5\std{0.5} & 0.269\std{0.000} & 0.467\std{0.022} & 200.748\std{0.822} \\

% ablation2_seed_0_TR_MP_WO
& & \ch & \ch & &
& {\bf 12.7}\std{0.3} & {\bf 0.361}\std{0.001} & 0.565\std{0.078} & 65.802\std{0.420} \\

% ablation2_seed_0_TR_ST_WO
& & \ch & & \ch &
& 15.0\std{0.3} & 0.356\std{0.002} & {\bf 0.418}\std{0.017} & 61.607\std{0.140} \\

\midrule

% Ours (ablation2_seed_0_TR_ST_W)
LayoutGAN++ & & \ch & & \ch & \ch
& 14.4\std{0.1} & 0.359\std{0.002} & 0.597\std{0.120} & {\bf 59.845}\std{0.595} \\

\bottomrule \end{tabular}
\end{center} \end{table*}

\begin{table*} \begin{center}
\caption{Ablation study on PubLayNet (Alignment 100x, Overlap 100x)}
\label{tab:ablation_study_publaynet}

\begin{tabular}{l cc cc c cccc} \toprule
& \multicolumn{2}{c}{Self-attention} & \multicolumn{2}{c}{Discriminator} & & \multicolumn{4}{c}{PubLayNet} \\
\cmidrule(lr){2-3} \cmidrule(lr){4-5} \cmidrule(lr){7-10}
Model & Stacked & Transformer & Max pool & Special token & Recon.
& FID $\downarrow$ & Max. IoU $\uparrow$ & Alignment $\downarrow$ & Overlap $\downarrow$ \\
\midrule

% ablation2_seed_0_SRLN_MP_WO
& \ch & & \ch & &
& 86.2\std{0.7} & 0.248\std{0.000} & 0.384\std{0.010} & 61.115\std{0.712} \\

% ablation2_seed_0_SRLN_MP_W
& \ch & & \ch & & \ch
& 44.5\std{0.3} & 0.349\std{0.001} & 0.448\std{0.003} & 31.833\std{0.372} \\

% ablation2_seed_0_SRLN_ST_WO
& \ch & & & \ch &
& 108.3\std{0.5} & 0.227\std{0.001} & 0.670\std{0.013} & 59.630\std{1.397} \\

% ablation2_seed_0_TR_MP_WO
& & \ch & \ch & &
& 48.8\std{0.3} & 0.302\std{0.000} & 0.294\std{0.008} & 29.682\std{0.194} \\

% ablation2_seed_0_TR_ST_WO
& & \ch & & \ch &
& 58.9\std{0.2} & 0.294\std{0.000} & {\bf 0.095}\std{0.002} & {\bf 17.257}\std{0.069} \\

\midrule

% Ours (ablation2_seed_0_TR_ST_W)
LayoutGAN++ & & \ch & & \ch & \ch
& {\bf 20.5}\std{0.3} & {\bf 0.357}\std{0.001} & 0.192\std{0.001} & 22.799\std{0.319} \\

\bottomrule \end{tabular}
\end{center} \end{table*}

\begin{table*} \begin{center}
\caption{Ablation study on Magazine (Alignment 100x, Overlap 100x)}
\label{tab:ablation_study_magazine}

\begin{tabular}{l cc cc c cccc} \toprule
& \multicolumn{2}{c}{Self-attention} & \multicolumn{2}{c}{Discriminator} & & \multicolumn{4}{c}{Magazine} \\
\cmidrule(lr){2-3} \cmidrule(lr){4-5} \cmidrule(lr){7-10}
Model & Stacked & Transformer & Max pool & Special token & Recon.
& FID $\downarrow$ & Max. IoU $\uparrow$ & Alignment $\downarrow$ & Overlap $\downarrow$ \\
\midrule

% ablation2_seed_0_SRLN_MP_WO
& \ch & & \ch & &
& 77.9\std{1.5} & 0.218\std{0.002} & 1.321\std{0.043} & 92.348\std{1.932} \\

% ablation2_seed_0_SRLN_MP_W
& \ch & & \ch & & \ch
& 44.0\std{0.8} & 0.228\std{0.002} & 1.010\std{0.036} & 81.331\std{1.311} \\

% ablation2_seed_0_SRLN_ST_WO
& \ch & & & \ch &
& 69.3\std{1.1} & 0.215\std{0.001} & 1.112\std{0.030} & 101.324\std{5.474} \\

% ablation2_seed_0_TR_MP_WO
& & \ch & \ch & &
& 13.8\std{0.4} & {\bf 0.260}\std{0.003} & 0.789\std{0.016} & 46.414\std{0.755} \\

% ablation2_seed_0_TR_ST_WO
& & \ch & & \ch &
& {\bf 13.3}\std{0.4} & 0.259\std{0.005} & {\bf 0.770}\std{0.021} & 39.405\std{0.682} \\

\midrule

% Ours (ablation2_seed_0_TR_ST_W)
LayoutGAN++ & & \ch & & \ch & \ch
& {\bf 13.3}\std{0.4} & 0.256\std{0.004} & 0.796\std{0.020} & {\bf 32.402}\std{0.887} \\

\bottomrule \end{tabular}
\end{center} \end{table*}
\setlength\tabcolsep{6pt}
}

%%% ULG figures
\newcommand{\ULGRowFig}[2]{
\includegraphics[width=13mm]{ulg/#1/real/#2_label.png} &
\frame{\includegraphics[width=13mm]{ulg/#1/layoutgan_wire/#2.png}} &
\frame{\includegraphics[width=13mm]{ulg/#1/layoutgan_rel/#2.png}} &
\frame{\includegraphics[width=13mm]{ulg/#1/ndn/#2.png}} &
\frame{\includegraphics[width=13mm]{ulg/#1/ours/#2.png}} &
\frame{\includegraphics[width=13mm]{ulg/#1/real/#2.png}}
}

\newcommand{\ULGRowFigNew}[3]{
\includegraphics[width=#3mm]{ulg/#1/real/#2_label.png} &
\frame{\includegraphics[width=13mm]{ulg/#1/layoutgan_wire/#2.png}} &
\frame{\includegraphics[width=13mm]{ulg/#1/layoutgan_rel/#2.png}} &
\frame{\includegraphics[width=13mm]{ulg/#1/ndn/#2.png}} &
\frame{\includegraphics[width=13mm]{ulg/#1/ours/#2.png}} &
\frame{\includegraphics[width=13mm]{ulg/#1/real/#2.png}}
}

\newcommand{\ULGFigure}{
\setlength\tabcolsep{0pt}
\begin{figure*}
\begin{center}
{\scriptsize
\begin{tabular}{c C{13.26mm}C{14mm}C{14mm}C{14mm}C{14mm}C{15mm} C{19mm}C{14mm}C{14mm}C{14mm}C{14mm}C{14mm}}
    & Label set & LayoutGAN-W~\cite{Li2019} & LayoutGAN-R~\cite{Li2019} & NDN-none~\cite{Lee2020} & LayoutGAN++ & Real data \hspace{1mm}
    & Label set & LayoutGAN-W~\cite{Li2019} & LayoutGAN-R~\cite{Li2019} & NDN-none~\cite{Lee2020} & LayoutGAN++ & Real data \\
    \rotatebox[origin=c]{90}{Rico} & \ULGRowFigNew{rico}{52}{11.01} \hspace{1mm} & \ULGRowFigNew{rico}{38}{14.27} \\
    \rotatebox[origin=c]{90}{PubLayNet} & \ULGRowFigNew{publaynet}{50}{10.07} \hspace{1mm} & \ULGRowFigNew{publaynet}{8}{10.13} \\
    \rotatebox[origin=c]{90}{Magazine} & \ULGRowFigNew{magazine}{1}{12.26} \hspace{1mm} & \ULGRowFigNew{magazine}{30}{18}\\
\end{tabular}}
\end{center}

\vspace*{-2mm}
\caption{Qualitative comparison of unconstrained layout generation. Label set
indicates the total number of labels and their type for each conditional
generation result. On the right we show the real data from which the label set
was taken.}
\label{fig:ulg_result}

\end{figure*}
\setlength\tabcolsep{6pt}
}

\newcommand{\ULGFigureSuppLoop}[3]{
\setlength\tabcolsep{0.5pt}
\begin{figure*}
\begin{center}
{\scriptsize
\begin{tabular}{C{12mm} C{13mm}C{13mm}C{13mm}C{13mm}C{13mm}C{14mm} }
    & Label set & LayoutGAN-W & LayoutGAN-R & NDN-none & LayoutGAN++ & Real data \\
    \forloop{index}{#2}{\value{index} < #3}{ \arabic{index} & \ULGRowFig{#1}{\arabic{index}} \\ }
\end{tabular}}
\end{center}

\caption{Qualitative comparison on unconstrained layout generation. (#1) }

\end{figure*}
\setlength\tabcolsep{6pt}
}

%%%% Constrained layout generation

%%% CLG (beautify setting) table
\newcommand{\CLGBTable}{
\setlength{\tabcolsep}{2.6pt}
\begin{table} \begin{center} 

\caption{Quantitative results with beautification constraints. Base model
refers to the unconstrained LayoutGAN++.  The values of Alignment and Overlap
are multiplied by $100\times$ for visibility.}
\label{tab:clg_beautify}
\vspace*{-2mm}

{\small \begin{tabular}{l cc cc cccc} \toprule
Model
& FID $\downarrow$ & Max. IoU $\uparrow$ & Alignment $\downarrow$ & Overlap $\downarrow$ \\
\midrule

Base model
& 20.48\std{0.29} & 0.36\std{0.00} & 0.19\std{0.00} & 22.80\std{0.32} \\                                            

CAL
& {\bf 13.31}\std{0.17} & {\bf 0.38}\std{0.00} & 0.16\std{0.00} & 14.27\std{0.19} \\

CLG-LO w/ Adam
& 21.79\std{0.38} & 0.36\std{0.00} & 0.16\std{0.00} & 1.18\std{0.04} \\

CLG-LO w/ CMA-ES
& 22.97\std{0.38} & 0.36\std{0.00} & {\bf 0.14}\std{0.00} & {\bf 0.02}\std{0.00} \\

\bottomrule \end{tabular} }
\end{center} \end{table}
\setlength\tabcolsep{6pt}
}

\newcommand{\CLGFig}[2]{
\frame{\includegraphics[width=\linewidth]{clg/#1/#2.png}}
}

\newcommand{\CLGBFigure}{
\setlength\tabcolsep{1pt}
\begin{figure}
\begin{center} \begin{tabular}{C{15mm} C{13mm}C{13mm}C{13mm}C{13mm}C{13mm}}
    Initial & \CLGFig{beautify}{init_6} & \CLGFig{beautify}{init_8} & \CLGFig{beautify}{init_12} & \CLGFig{beautify}{init_18} & \CLGFig{beautify}{init_31} \\
    Optimized & \CLGFig{beautify}{opt_6} & \CLGFig{beautify}{opt_8} & \CLGFig{beautify}{opt_12} & \CLGFig{beautify}{opt_18} & \CLGFig{beautify}{opt_31} \\
    % Real data & \CLGFig{beautify}{gt_6} & \CLGFig{beautify}{gt_8} & \CLGFig{beautify}{gt_12} & \CLGFig{beautify}{gt_18} & \CLGFig{beautify}{gt_31}
\end{tabular} \end{center}

\vspace*{-2mm}
\caption{Qualitative results with beautification constraints for CLG-LO w/
CMA-ES. Initial unconditioned generation results are shown in the top row and
the optimized results are shown in the bottom row.}
\label{fig:clg_beautify}

\end{figure}
\setlength\tabcolsep{6pt}
}

\newcommand{\CLGRelFig}[4]{
\includegraphics[width=#2mm]{clg/relation/rico_#1/relation.png} &
\frame{\includegraphics[width=13mm]{clg/relation/rico_#1/init.png}} &
\frame{\includegraphics[width=13mm]{clg/relation/rico_#1/opt.png}} &
&
\includegraphics[width=#3mm]{clg/relation/publaynet_#1/relation.png} &
\frame{\includegraphics[width=13mm]{clg/relation/publaynet_#1/init.png}} &
\frame{\includegraphics[width=13mm]{clg/relation/publaynet_#1/opt.png}} &
&
\includegraphics[width=#4mm]{clg/relation/magazine_#1/relation.png} &
\frame{\includegraphics[width=13mm]{clg/relation/magazine_#1/init.png}} &
\frame{\includegraphics[width=13mm]{clg/relation/magazine_#1/opt.png}}
}

\newcommand{\CLGRFigure}{
\setlength\tabcolsep{0.5pt}
\begin{figure*}
\small{ \begin{center} \begin{tabular}{L{30mm}C{13.5mm}C{13.5mm} C{2mm} L{25.16mm}C{13.5mm}C{13.5mm} C{2mm} L{24mm}C{13.5mm}C{13.5mm}}
    \multicolumn{3}{c}{Rico} & & \multicolumn{3}{c}{PubLayNet} & & \multicolumn{3}{c}{Magazine} \\
    \multicolumn{1}{c}{Constraints} & Initial & Optimized & &
    \multicolumn{1}{c}{Constraints} & Initial & Optimized & &
    \multicolumn{1}{c}{Constraints} & Initial & Optimized \\
    \CLGRelFig{1}{30}{23.68}{24} \\
    \CLGRelFig{2}{24.63}{25.16}{24} \\
\end{tabular} \end{center} }

\vspace*{-2mm}
\caption{Qualitative results with relational constraints for the three datasets
for our prposed CLG-LO w/ CMA-ES. In each column, for each result we show the
constraints on the left, the initial unconstrained generation result in the
middle, and the optimized result on the right. }
\label{fig:clg_relation}

\end{figure*}
\setlength\tabcolsep{6pt}
}

\newcommand{\CLGRPlot}{
\begin{figure*}
\begin{center}
\includegraphics[width=\linewidth]{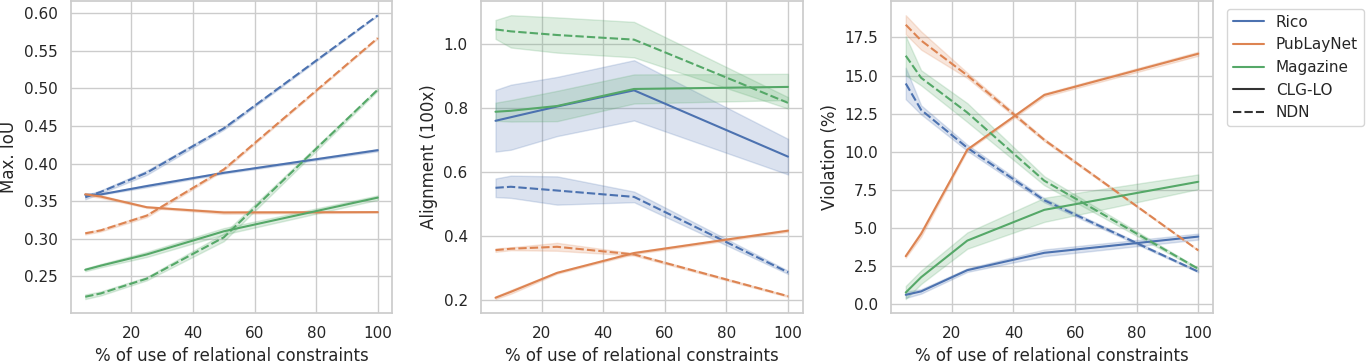}
\end{center}

\vspace*{-2mm}
\caption{Quantitative results with relational constraints. The different colors
correspond to each of the three datasets. The solid lines denotes CLG-LO, and
the dashed lines denotes NDN. Higher is better for Max.~IoU, and lower is
better for Alignment and Violation. Our proposed CLG-LO approach often
outperforms NDN when only a small part of relations is specified.}
\label{fig:clg_plot}
\end{figure*}
}

%%% CLG table
\newcommand{\CLGRTable}{
\setlength{\tabcolsep}{2.6pt}
\begin{table*} \begin{center} 
\caption{Quantitative results with relational constraints when 10\% of all the
relational constraints are used.  The values of Alignment are multiplied by
$100\times$ for visibility.}
\label{tab:clg_relation}

\footnotesize{ \begin{tabular}{l ccc ccc ccc} \toprule
\multicolumn{1}{r}{Dataset} & \multicolumn{3}{c}{Rico} & \multicolumn{3}{c}{PubLayNet} & \multicolumn{3}{c}{Magazine} \\
\cmidrule(lr){2-4} \cmidrule(lr){5-7} \cmidrule(lr){8-10}
Model
& Max. IoU $\uparrow$ & Alignment $\downarrow$ & Const. violation (\%) $\downarrow$
& Max. IoU $\uparrow$ & Alignment $\downarrow$ & Const. violation (\%) $\downarrow$
& Max. IoU $\uparrow$ & Alignment $\downarrow$ & Const. violation (\%) $\downarrow$ \\
\midrule

NDN \cite{Lee2020}
& {\bf 0.36}\std{0.00} & {\bf 0.56}\std{0.03} & 12.75\std{0.27}
& 0.31\std{0.00} & 0.36\std{0.00} & 17.30\std{0.54}
& 0.23\std{0.00} & 1.04\std{0.05} & 14.85\std{0.44}
\\

CLG-LO
& {\bf 0.36}\std{0.00} & 0.77\std{0.09} & {\bf 0.84}\std{0.13}
& {\bf 0.36}\std{0.00} & {\bf 0.23}\std{0.01} & {\bf 4.61}\std{0.17}
& {\bf 0.26}\std{0.00} & {\bf 0.79}\std{0.03} & {\bf 1.77}\std{0.39}
\\

\bottomrule \end{tabular} }
\end{center} \end{table*}
\setlength{\tabcolsep}{6pt}
}

\newcommand{\CLGRTableOld}{
\setlength{\tabcolsep}{2.6pt}

\begin{table*} \begin{center} 
\caption{Constrained layout generation results on Rico. (Alignment 100x, Overlap 100x)}
\label{tab:clg_ndn_rico}

\begin{tabular}{cl cccccc} \toprule
\# const. & Model
& FID $\downarrow$ & Max. IoU $\uparrow$ & Alignment $\downarrow$ & Overlap $\downarrow$
& $c_\mathrm{size}$ violation (\%) $\downarrow$ & $c_\mathrm{loc}$ violation (\%) $\downarrow$ \\

\midrule \multirow{2}{*}{0\%}
& NDN-none \cite{Lee2020}
& {\bf 13.8}\std{0.3} & 0.352\std{0.001} & {\bf 0.557}\std{0.029} & {\bf 54.753}\std{0.286} & - & - \\

& LayoutGAN++
& 14.4\std{0.1} & {\bf 0.359}\std{0.002} & 0.597\std{0.120} & 59.845\std{0.595} & - & - \\

\midrule \multirow{2}{*}{5\%}
& NDN \cite{Lee2020}
& {\bf 13.4}\std{0.3} & 0.355\std{0.002} & {\bf 0.552}\std{0.026} & {\bf 54.898}\std{0.367} & 11.36\std{0.60} & 17.60\std{1.54} \\

& LayoutGAN++
& 13.5\std{0.2} & {\bf 0.358}\std{0.001} & 0.761\std{0.086} & 59.527\std{0.314} & {\bf 0.59}\std{0.23} & {\bf 0.64}\std{0.15} \\

\midrule \multirow{2}{*}{10\%}
& NDN \cite{Lee2020}
& 12.9\std{0.2} & {\bf 0.362}\std{0.002} & {\bf 0.555}\std{0.031} & {\bf 54.869}\std{0.374} & 9.92\std{0.56} & 15.57\std{0.22} \\

& LayoutGAN++
& {\bf 12.6}\std{0.1} & 0.359\std{0.000} & 0.773\std{0.091} & 58.369\std{0.405} & {\bf 0.84}\std{0.19} & {\bf 0.84}\std{0.09} \\

\midrule \multirow{2}{*}{25\%}
& NDN \cite{Lee2020}
& {\bf 10.5}\std{0.2} & {\bf 0.388}\std{0.003} & {\bf 0.543}\std{0.039} & {\bf 54.596}\std{0.387} & 8.30\std{0.19} & 12.22\std{0.29} \\

& LayoutGAN++
& 12.5\std{0.1} & 0.370\std{0.001} & 0.806\std{0.083} & 56.929\std{0.255} & {\bf 2.44}\std{0.15} & {\bf 2.02}\std{0.07} \\

\midrule \multirow{2}{*}{50\%}
& NDN \cite{Lee2020}
& {\bf 6.6}\std{0.2} & {\bf 0.447}\std{0.002} & {\bf 0.524}\std{0.015} & 54.829\std{0.351} & 5.79\std{0.13} & 7.85\std{0.27} \\

& LayoutGAN++
& 12.7\std{0.2} & 0.388\std{0.001} & 0.856\std{0.084} & {\bf 53.977}\std{0.181} & {\bf 4.04}\std{0.23} & {\bf 2.70}\std{0.23} \\

\midrule \multirow{2}{*}{100\%}
& NDN-all \cite{Lee2020}
& {\bf 2.3}\std{0.0} & {\bf 0.597}\std{0.000} & {\bf 0.288}\std{0.005} & 53.168\std{0.042} & {\bf 2.06}\std{0.05} & {\bf 2.21}\std{0.04} \\

& LayoutGAN++
& 12.7\std{0.3} & 0.418\std{0.001} & 0.649\std{0.050} & {\bf 48.951}\std{0.236} & 5.51\std{0.14} & 3.35\std{0.25} \\

\bottomrule \end{tabular}
\end{center} \end{table*}

\begin{table*} \begin{center} 
\caption{Constrained layout generation results on PubLayNet. (Alignment 100x, Overlap 100x)}
\label{tab:clg_ndn_publaynet}

\begin{tabular}{cl cccccc} \toprule
\# const. & Model
& FID $\downarrow$ & Max. IoU $\uparrow$ & Alignment $\downarrow$ & Overlap $\downarrow$
& $c_\mathrm{size}$ violation (\%) $\downarrow$ & $c_\mathrm{loc}$ violation (\%) $\downarrow$ \\

\midrule \multirow{2}{*}{0\%}
& NDN-none \cite{Lee2020}
& 35.7\std{0.3} & 0.305\std{0.002} & 0.355\std{0.008} & {\bf 16.499}\std{0.287} & - & - \\

& LayoutGAN++
& {\bf 20.5}\std{0.3} & {\bf 0.357}\std{0.001} & {\bf 0.192}\std{0.001} & 22.799\std{0.319} & - & - \\

\midrule \multirow{2}{*}{5\%}
& NDN \cite{Lee2020}
& 35.9\std{0.5} & 0.307\std{0.001} & 0.357\std{0.005} & {\bf 16.985}\std{0.294} & 12.05\std{0.41} & 24.62\std{0.80} \\

& LayoutGAN++
& {\bf 17.9}\std{0.4} & {\bf 0.359}\std{0.002} & {\bf 0.208}\std{0.004} & 22.475\std{0.180} & {\bf 2.64}\std{0.06} & {\bf 3.66}\std{0.13} \\

\midrule \multirow{2}{*}{10\%}
& NDN \cite{Lee2020}
& 35.8\std{0.6} & 0.311\std{0.001} & 0.362\std{0.004} & {\bf 17.538}\std{0.241} & 11.47\std{0.51} & 23.12\std{0.58} \\

& LayoutGAN++
& {\bf 18.3}\std{0.3} & {\bf 0.356}\std{0.002} & {\bf 0.227}\std{0.005} & 23.330\std{0.179} & {\bf 4.42}\std{0.15} & {\bf 4.80}\std{0.22} \\

\midrule \multirow{2}{*}{25\%}
& NDN \cite{Lee2020}
& 32.9\std{0.5} & {\bf 0.331}\std{0.002} & 0.368\std{0.011} & {\bf 17.998}\std{0.284} & {\bf 9.79}\std{0.09} & 20.23\std{0.33} \\

& LayoutGAN++
& {\bf 23.6}\std{0.2} & 0.342\std{0.001} & {\bf 0.286}\std{0.003} & 23.909\std{0.220} & 10.81\std{0.07} & {\bf 9.47}\std{0.11} \\

\midrule \multirow{2}{*}{50\%}
& NDN \cite{Lee2020}
& {\bf 22.5}\std{0.4} & {\bf 0.392}\std{0.001} & {\bf 0.344}\std{0.003} & {\bf 15.919}\std{0.156} & {\bf 6.87}\std{0.08} & 14.69\std{0.14} \\

& LayoutGAN++
& 29.6\std{0.4} & 0.335\std{0.002} & 0.348\std{0.003} & 22.300\std{0.125} & 14.20\std{0.19} & {\bf 13.25}\std{0.06} \\

\midrule \multirow{2}{*}{100\%}
& NDN-all \cite{Lee2020}
& {\bf 4.5}\std{0.1} & {\bf 0.567}\std{0.001} & {\bf 0.213}\std{0.002} & {\bf 4.056}\std{0.044} & {\bf 1.54}\std{0.03} & {\bf 5.52}\std{0.05} \\

& LayoutGAN++
& 34.5\std{0.5} & 0.335\std{0.001} & 0.418\std{0.003} & 17.238\std{0.205} & 16.50\std{0.14} & 16.36\std{0.12} \\

\bottomrule \end{tabular}
\end{center} \end{table*}

\begin{table*} \begin{center} 
\caption{Constrained layout generation results on Magazine. (Alignment 100x, Overlap 100x)}
\label{tab:clg_ndn_magazine}

\begin{tabular}{cl cccccc} \toprule
\# const. & Model
& FID $\downarrow$ & Max. IoU $\uparrow$ & Alignment $\downarrow$ & Overlap $\downarrow$
& $c_\mathrm{size}$ violation (\%) $\downarrow$ & $c_\mathrm{loc}$ violation (\%) $\downarrow$ \\

\midrule \multirow{2}{*}{0\%}
& NDN-none \cite{Lee2020}
& 23.3\std{0.9} & 0.220\std{0.002} & 1.048\std{0.026} & {\bf 30.311}\std{0.774} & - & - \\

& LayoutGAN++
& {\bf 13.3}\std{0.4} & {\bf 0.256}\std{0.004} & {\bf 0.796}\std{0.020} & 32.402\std{0.887} & - & - \\

\midrule \multirow{2}{*}{5\%}
& NDN \cite{Lee2020}
& 22.6\std{0.8} & 0.223\std{0.003} & 1.047\std{0.026} & {\bf 30.546}\std{0.476} & 10.27\std{0.83} & 22.33\std{1.81} \\

& LayoutGAN++
& {\bf 13.2}\std{0.7} & {\bf 0.259}\std{0.002} & {\bf 0.790}\std{0.025} & 31.490\std{0.596} & {\bf 0.56}\std{0.33} & {\bf 0.97}\std{0.47} \\

\midrule \multirow{2}{*}{10\%}
& NDN \cite{Lee2020}
& 22.1\std{0.4} & 0.227\std{0.002} & 1.041\std{0.045} & {\bf 30.345}\std{0.748} & 9.15\std{0.93} & 20.55\std{0.80} \\

& LayoutGAN++
& {\bf 13.7}\std{0.2} & {\bf 0.264}\std{0.002} & {\bf 0.793}\std{0.030} & 31.052\std{0.794} & {\bf 1.97}\std{0.32} & {\bf 1.57}\std{0.47} \\

\midrule \multirow{2}{*}{25\%}
& NDN \cite{Lee2020}
& 19.8\std{0.3} & 0.247\std{0.002} & 1.029\std{0.049} & 31.076\std{0.771} & 8.37\std{0.77} & 16.77\std{0.58} \\

& LayoutGAN++
& {\bf 13.8}\std{0.4} & {\bf 0.279}\std{0.003} & {\bf 0.808}\std{0.042} & {\bf 31.060}\std{0.640} & {\bf 5.26}\std{0.41} & {\bf 3.08}\std{0.55} \\

\midrule \multirow{2}{*}{50\%}
& NDN \cite{Lee2020}
& 13.3\std{0.2} & 0.302\std{0.005} & 1.015\std{0.050} & 30.620\std{0.359} & {\bf 6.06}\std{0.17} & 10.14\std{0.40} \\

& LayoutGAN++
& {\bf 12.4}\std{0.5} & {\bf 0.310}\std{0.004} & {\bf 0.861}\std{0.041} & {\bf 30.057}\std{0.766} & 7.74\std{0.73} & {\bf 4.64}\std{0.71} \\

\midrule \multirow{2}{*}{100\%}
& NDN-all \cite{Lee2020}
& {\bf 5.4}\std{0.1} & {\bf 0.498}\std{0.001} & {\bf 0.818}\std{0.016} & {\bf 26.992}\std{0.087} & {\bf 2.75}\std{0.09} & {\bf 1.89}\std{0.06} \\

& LayoutGAN++
& 11.0\std{0.4} & 0.355\std{0.003} & 0.867\std{0.037} & 27.053\std{0.404} & 10.30\std{0.39} & 5.77\std{0.54} \\

\bottomrule \end{tabular}
\end{center} \end{table*}
\setlength\tabcolsep{6pt}
}

%% file: intro.tex
\section{Introduction}

% Our daily communication through visual media such as documents, magazines, and user interfaces is essential,
% and they are graphically designed to make the recipient's response what the sender intended.
% Layout, the arrangement of the size and position of the elements to be displayed, is one of the major parts of graphic design,
% which aims to control the semantic relationship, priority, and reading order among elements while respecting visual aesthetics.
% A less experienced designer will go through a trial-and-error process to find a layout that satisfies many design constraints simultaneously.
% The constraints can be internal, derived from the one's design experience and preference, or external, such as visual media regulations and client requirements.
% An automatic search of plausible layout candidates that satisfy the constraints can aid the design process.
Visual media contents are organized using design layouts to facilitate the
conveying of information.
Design layout consists of the arrangement of the size and position of the
elements to be displayed, and is a critical part of graphic design.
In general, articles start with a text title, followed by headings and the main
text, usually in a top to bottom order.
Mobile user interfaces arrange navigation, images, texts, or buttons cleanly in
a given display resolution with fluid layouts.
The semantic relationships, priority, and reading order of elements is
carefully decided by graphic designers while considering the overall visual
aesthetics of the design.
Inexperienced designers often face the difficulty of producing high-quality
presentations while conveying the designated message and maintaining
fundamental design considerations such as alignment or overlap.
Design constraints can be internal, derived from the one's design experience
and preference, or external, such as visual media regulations and client
requirements.
Automatic search of plausible layout candidates, such as we propose in this
paper, can greatly aid in the design process.

Several attempts have been made to automatically generate graphic layouts in the
computer graphics community~\cite{ODonovan2014,ODonovan2015}.  Recent
studies~\cite{Li2019,Jyothi2019,Arroyo2021} using unconstrained deep generative
models have shown to be able to generate plausible layouts thanks to large
scale datasets of design examples.  Some work explicitly introduce design
constraints like alignment or overlap avoidance by additional losses or
conditioning~\cite{Li2020,Lee2020}.  However, one drawback of integrating
constraints in the learning objective is that a model must be fit to a new
condition or a new loss when there appears a new constraint a user wishes to
incorporate. We instead opt to perform the optimization in the latent space of
the generative model, being complementary to and allowing for the usage of
existing off-the-shelf models.
% This is undesirable, given that constraints could change drastically during the design process
% due to internal factors such as changes in the designer's preferences
% and external factors such as changes in the client's requirements.

In this work, we propose a novel framework, which we call Constrained Layout Generation via
Latent Optimization~(CLG-LO), that defines constrained layout generation as a
constrained optimization problem in the latent space of the model.
An overview of the proposed
framework is illustrated in Fig.~\ref{fig:teaser}.
In our approach, we use a
Generative Adversarial Network (GAN) trained in the unconstrained setting and
model user specifications as a constrained optimization program.
We optimize the latent code of the unconstrained model with an iterative algorithm to
find a layout that satisfies the specified constraints.  Our
framework allows the user to use a single pre-trained GAN and incorporate various
constraints into the layout generation as needed, eliminating the
computationally expensive need of re-training of the model.

Although our approach can work with off-the-shelf generative layout models, in
addition to CLG-LO framework, we also propose a
Transformer~\cite{Vaswani2017} based layout GAN model, which we name
LayoutGAN++.  Relationships between elements can be well captured by
Transformers in both the generator and the discriminator.  With the help of
representation learning of the discriminator through auxiliary layout
reconstruction~\cite{Liu2021}, LayoutGAN++ significantly improves the
performance of the LayoutGAN~\cite{Li2019} for unconstrained layout generation.

We validate our proposed methods using three public datasets of graphic
layouts.  We design two constrained generation settings similar to real use
cases. In the unconstrained generation task, LayoutGAN++ obtains comparable or
better results than the existing methods.  Using LayoutGAN++ as the backend
model, CLG-LO shows significant improvements in the constrained generation task.

We summarize our contributions as follows:
\begin{itemize}
    \item A framework to generate layouts that satisfies given constraints by optimizing latent codes.
    \item An architecture and methodology for layout GAN that allows for stable training and generation of high-quality \del{design }layouts.
    \item Extensive experiments and state-of-the-art results using public datasets for unconstrained and constrained layout generation.
\end{itemize}

%% file: related.tex
\section{Related Work}

\subsection{Layout Generation}
There has been several studies on generating layout, both with or without user
specification.  Classic optimization
approaches~\cite{ODonovan2014,ODonovan2015} manually designed energy functions
with a large number of constraints that a layout should satisfy.
Recent works have utilized neural networks to learn a generative model of layout.  LayoutVAE
trained two types of Variational Auto-Encoders (VAE) to generate bounding boxes to the given label
set~\cite{Jyothi2019}.  LayoutGAN trained relational generator by employing a wireframe
renderer that rasterize bounding boxes and allows for training with a pixel-based
discriminator~\cite{Li2019}.  Later, LayoutGAN was extended to include attribute
conditioning~\cite{Li2020}.  \citet{Zheng2019} reported a raster layout
generator conditioned on the given images, keywords, and attributes.
READ~\cite{Patil2020} trained a hierarchical auto-encoder to generate document
layout structures. \citet{Lee2020} proposed graph-based networks called
Neural Design Networks (NDN) that explicitly infer element relations from
partial user specification. Very recently, \citet{Gupta2021} described a
Transformer-based model to generate layout in various domains.  Also,
\citet{Arroyo2021} reported a VAE model that
generated layouts using self-attention networks.
Apart from graphic design layouts, there has also been research on generating
indoor scene layouts~\cite{Henderson2017,Ritchie2019,Zhang2020deep}.

Our work considers both unconstrained generation~\cite{Gupta2021,Arroyo2021}
and constrained generation~\cite{Li2020,Lee2020}.  We build our unconstrained
layout generator based on LayoutGAN~\cite{Li2019}, and apply user layout
specification as constraints to a learned generator. Unlike NDN~\cite{Lee2020},
we only need a single model to generate constrained layouts.

% \subsection{Generative Adversarial Networks (GAN) and Transformer}

% \subsection{Constrained Layout Optimization}
% \begin{itemize}
%     \item Grids~\cite{Dayama2020}
%     \item Scout
%     \item Content-aware layout generation
%     \item Attribute-conditioned LayoutGAN~\cite{Li2020}
%     \item Neural Design Networks~\cite{Lee2020}
% \end{itemize}

% TODO: Maybe interacitve layout systems?

% \subsection{Latent Space Exploration}
% % \begin{itemize}
% %     \item Intermediate Layer Optimization
% %     \item PULSE~\cite{Menon2020}
% %     \item DSS~\cite{Chiu2020}
% % \end{itemize}

% For constrained layout generation, we get inspirations from a few recent work
% on latent space exploration~\cite{Menon2020,Chiu2020}.  In PULSE,
% \citet{Menon2020} search through the latent space of high-resolution
% facial photos to achieve super-resolution of low-quality photos.  In DSS,
% \citet{Chiu2020} utilize a learned generative model to navigate through
% latent space in an interactive system.  Our layout generation
% approach shares the concept of latent space exploration, and we seek to
% find a latent representation of layout such that the resulting layout satisfies
% user-specified constraints.

\subsection{Latent Space Exploitation}
With the recent progress in image synthesis using deep generative models~\cite{Karras2018,Karras2019},
much of the research utilizing the latent space have been made in the image domain.
In real image editing,
the mainstream research involves projecting the target image into the latent space
and performing non-trivial image editing with user input on the learned manifold~\cite{Zhu2016,Bau2019,Zhu2020}.
\citet{DGP} also used the natural image priors learned by GAN
and applied them to various image restoration tasks such as inpainting and colorization in a unified way.
\citet{Menon2020} search through the latent space of high-resolution
facial photos to achieve super-resolution of low-quality photos.

The utilization of latent variables in deep generative models have been less
studied in non-image domains.  \citet{Umetani2017} proposed an interactive
interface that uses a learned auto-encoder to find the shape of a 3D model by
adjusting latent variables.  \citet{Schrum2020} proposed an interface
consisting of interactive evolutionary search and direct manipulation of latent
variables for the game level design.  \citet{Chiu2020} proposed a method to
efficiently explore latent space in a human-in-the-loop fashion using a learned
generative model, and validated it in the tasks of generating images, sounds,
and 3D models.

Our layout generation approach shares the concept of latent space exploration,
and we seek to find a latent representation of layout such that the resulting
layout satisfies user-specified constraints.

%% file: approach.tex
\section{Approach}

\begin{figure*}
  \begin{center}
    \includegraphics[width=\linewidth]{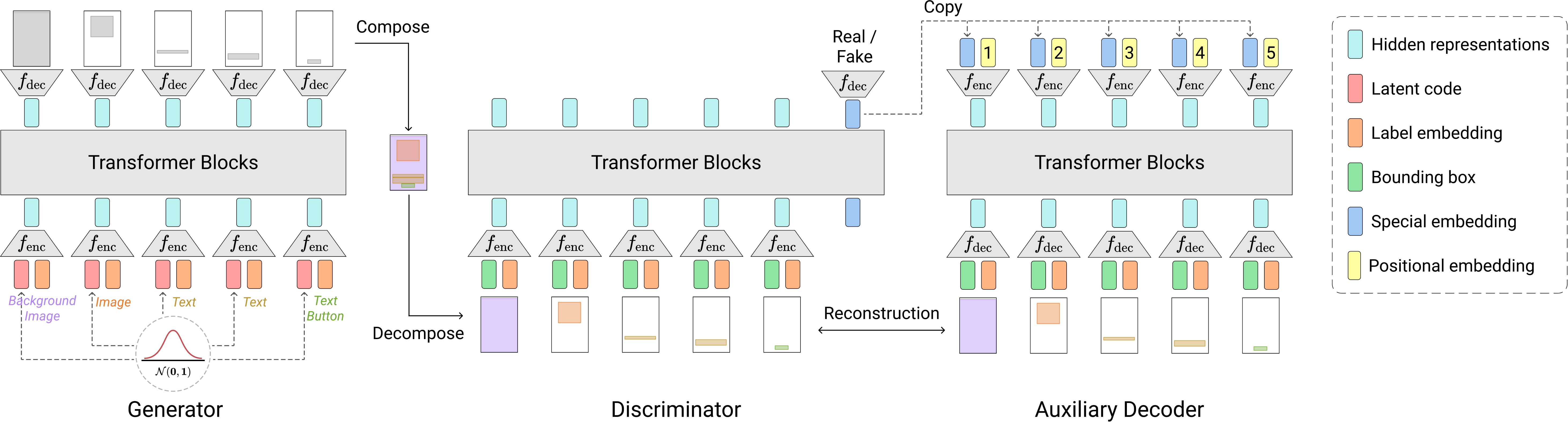}
  \end{center}
  \vspace*{-2mm}
  \caption{Overview of our proposed LayoutGAN++ model.}
  \label{fig:model}
\end{figure*}

Our goal is to generate a semantically plausible and high-quality design layout from a
set of element labels and constraints specified by the user.  We first
train an unconstrained generative model of layout denoted \textit{LayoutGAN++},
and later utilize the model for constrained generation tasks.

\subsection{LayoutGAN++}

In unconstrained generation, we take a set of elements and assign size and
location to each element.  We follow LayoutGAN~\cite{Li2019} and formulate our
model, which we refer \emph{LayoutGAN++}, in the following.  Formally, our
generator \del{$G(Z \;|\; L)$}\add{$G \colon (Z, L) \mapsto B$} takes a set of
randomly-generated codes $Z=\{ \mathbf{z}_i
\}_{i=1}^N$ and a conditional multiset of labels $L = \{\!\!\{ l_i
\}\!\!\}_{i=1}^N$ as input, and outputs a set of bounding boxes $B=\{
\mathbf{b}_i \}_{i=1}^N$, where $\mathbf{b}_i \in [0, 1]^4$ represents the
position and size of the element in normalized coordinates.  $N$ is the
number of elements in a layout, and the subscript $i$ in $Z$, $L$, and $B$
refers to the same $i$-th element.  The definition of a label $l$ depends on
the dataset; \eg, text or table elements in PubLayNet dataset.  Our
discriminator \del{$D(B \;|\; L)$}\add{$D \colon (B, L) \mapsto r \in [0,1]$} takes the generated bounding boxes $B$ and
conditional labels $L$ as input, and outputs a scalar value which quantifies
the realism of layout, as well as attempts at reconstructing the given
bounding boxes from the internal representation.  We show in
Fig.~\ref{fig:model} the overall architecture of our model.

\subsubsection{Generator}
Our generator consists of the following:
\begin{align}
    \mathbf{z}_i &\sim \mathcal{N}(\mathbf{0},\mathbf{I}), \\
    \mathbf{h}_i &= f_\mathrm{enc}(\mathbf{z}_i, l_i; \theta), \\
    \{\mathbf{h}_i'\} &= \mathrm{Transformer}(\{\mathbf{h}_i\}; \theta), \\
    \mathbf{b}_i &= f_\mathrm{dec}(\mathbf{h}_i'; \theta),
\end{align}
\noindent where $f_\mathrm{enc}$, $f_\mathrm{dec}$ are multi-layer perceptrons,
$\mathbf{h}_i$ and $\mathbf{h}_i'$ are hidden representations for each element,
and $\theta$ is the parameters for the generator.  We adopt the
Transformer block~\cite{Vaswani2017} to learn relational representation among
elements, in contrast to LayoutGAN~\cite{Wang2018} that utilizes a dot
product-based non-local block with a residual connection.

\subsubsection{Discriminator}
Our discriminator has a similar architecture to our generator.
\begin{align}
    \mathbf{h}_i &= f_\mathrm{enc}(\mathbf{b}_i, l_i; \phi), \\
    \mathbf{h}_\mathrm{const}' &= \mathrm{Transformer}(\mathbf{h}_\mathrm{const}, \{\mathbf{h}_i\}; \phi), \\
    y &= f_\mathrm{dec}(\mathbf{h}_\mathrm{const}'; \phi),
\end{align}
\noindent where $\mathbf{h}_\mathrm{const}$ is a special learnable embedding
appended to the hidden element representations, $\mathbf{h}_\mathrm{const}'$ is
the corresponding output for the learnable embedding after the Transformer
block, $y$ is the quantity to evaluate the reality of the given input, and
$\phi$ is the parameters of the discriminator.  We do not employ the wireframe
renderer of LayoutGAN~\cite{Wang2018}, because we find that the raster domain
discriminator becomes unstable with limited dataset size. We compare with
LayoutGAN in our experiments.

\subsubsection{Auxiliary Decoder}\label{sec:auxiliary_decoder}
We empirically find that in well-aligned layout domains such as documents, the
discriminator is trained to be sensitive to alignment and less sensitive to
positional trends, \add{\ie, it only cares if the elements are aligned, and does not
care about unusual layouts such as placing the header element at the bottom.}
Following the self-supervised learning of \citet{Liu2021},
we apply additional regularization to the discriminator so that the
discriminator becomes aware of the positional trends. We add an auxiliary
decoder to reconstruct the bounding boxes given to the discriminator from the
internal representation $\mathbf{h}_\mathrm{const}'$:
\begin{align}
    \mathbf{h}_i &= f_\mathrm{enc}(\mathbf{h}_\mathrm{const}', \mathbf{p}_i; \xi), \\
    \{\mathbf{h}_i'\} &= \mathrm{Transformer}(\{\mathbf{h}_i\}; \xi), \\
    \hat{\mathbf{b}}_i, \hat{l}_i &= f_\mathrm{dec}(\mathbf{h}_i'; \xi),
\end{align}
\noindent where $\mathbf{p}_i$ is a \add{learnable} positional embedding
\add{initialized with the uniform distribution of $[0, 1]$}, $\hat{\mathbf{b}}_i
\in \hat{B}$ is a reconstructed bounding box, $\hat{l}_i \in \hat{L}$ is a
reconstructed label, and $\xi$ is the parameters of the auxiliary decoder.

\subsubsection{Training objective}
The objective function of our model is the following:
% TODO: Describe the loss/objective function here.
\begin{align}
  \min_{\theta} \max_{\phi, \xi} ~
  & E_{(B, L) \sim P_\mathrm{data}}\left[ D(B \del{\;|\;}\add{,} L; \phi) - \mathcal{L}_\mathrm{rec}\left(B, L, \hat{B}(\phi, \xi), \hat{L}(\phi, \xi)\right) \right] + \nonumber \\
  & E_{Z \sim \mathcal{N}, L \sim P_\mathrm{data}}\big[ 1 - D\big( G(Z\del{\;|\;}\add{,} L; \theta) \del{\;\big|\;}\add{,} L; \phi\big) \big]
\end{align}
\noindent where we denote the reconstruction loss by $\mathcal{L}_\mathrm{rec}$.
The reconstruction loss measures the similarity between two sets of bounding
boxes and labels, and we employ mean squared error for bounding boxes,
and cross entropy for labels.  We compute the reconstruction loss by
first sorting the bounding boxes in lexicographic order of the ground-truth
positions~\cite{Carlier2020}.
% These modifications are verified in a latter ablation study.

%%%%%%%%%%%%%%%%%%%%%%%%%%%%%%%%%%%%%%%%%%%%%%%%%%%%%%%%%%%%%%%%%%%%%%%%%%%
\subsection{Constrained Layout Generation via Latent Optimization (CLG-LO)}
Let us consider when there are user-specified constraints, such as \emph{an element A must be above an element B}.
From the perspective of the generator, such constraints restricts the available output space.
We formulate the generation with user specification in a constrained optimization problem.
Given a pre-trained generator \del{$\bar{G}$}\add{$\hat{G}$} and discriminator \del{$\bar{D}$}\add{$\hat{D}$}, and a set of constraints $C$,
we define the constrained minimization problem regarding latent codes $Z$:
\begin{eqnarray}
  \min_Z & \del{-\bar{D}\left(\bar{G}(Z \;|\; L) \;\big|\; L\right)}\add{-\hat{D}\big(\hat{G}(Z, L), L\big)} \nonumber & \\
  \mathrm{s.t.} &  c_n\big( \del{\bar{G}(Z \;|\; L)}\add{\hat{G}(Z, L)} \big) = 0, & n=1,\ldots,|C|.
  \label{eqn:const}
\end{eqnarray}
The intuition is that we seek to find bounding boxes that looks as realistic as possible to the discriminator and satisfies the user-specified constraints.
Once the optimal latent codes $Z^\ast$ is found, we can obtain bounding boxes $B^\ast$ that satisfy the constraints as follows:
\begin{equation}
  B^\ast = \del{\bar{G}(Z^\ast \;|\; L)}\add{\hat{G}(Z^\ast, L)}.
\end{equation}

We use the augmented Lagrangian method~\cite{Nocedal2006}, which is one of the
widely used algorithms for solving nonlinear optimization problems.  In this
method, the constrained problem is transformed into an unconstrained problem
that optimizes the augmented Lagrangian function, which combines the Lagrangian
and penalty functions.  Let us rewrite 
$f(Z) = \del{-\bar{D}\left(\bar{G}(Z \;|\; L) \;\big|\; L\right)\,}\add{-\hat{D}\big(\hat{G}(Z, L), L\big)}$
and $h_n(Z) = c_n\big( \del{\bar{G}(Z \;|\; L)}\add{\hat{G}(Z, L)} \big)$
in Eq.~\eqref{eqn:const} for brevity,
then we define the following augmented Lagrangian function~$L_A$,
\begin{equation}
  L_A(Z\del{, \mathbf{\lambda}}; \add{\mathbf{\lambda},} \mu) \,=\, f(Z)
  \,+\, \sum_{n=1}^{|C|} \mathbf{\lambda}_n h_n(Z)
  \,+\, \frac{\mu}{2} \sum_{n=1}^{|C|} h_n(Z)^2,
  \label{eqn:auglag}
\end{equation}
\noindent where $\mathbf{\lambda}$ are the Lagrange multipliers and $\mu > 0$
is a penalty parameter to weight the quadratic functions.

In this method, the Lagrange multipliers are updated according to the extent of
constraint violation, and the penalty parameter is gradually increased to make
the impact of the constraints larger.  Let $k$ be the current iteration, the
update equations are expressed as:
\begin{align}
  \mathbf{\lambda}^{k+1}_n &= \mathbf{\lambda}^{k}_n + \mu_k h_n(Z_k) \label{eqn:update_lag} \\
  \mu_{k+1} &= \alpha \mu_{k} \label{eqn:update_mu},
\end{align}
\noindent where $\alpha$ is a predefined hyperparameter.

Algorithm~\ref{algo:generate_layout} summarizes the procedure of our method.
We repeat the main loop until the amount of constraint violation is sufficiently small
or the iteration count reaches the maximum number of iterations $k_\mathrm{max}$.
We set $\alpha=3$, $\mu_0=1$, $\mathbf{\lambda}^0=\mathbf{0}$, and $k_\mathrm{max}=5$ in the experiments.
For the inner optimizer,
we use either Adam~\cite{Kingma2015} with a learning rate of 0.01
or CMA-ES~\cite{Hansen2016} with a initial sigma value of 0.25, and both run for 200 iterations.
We compare in Sec~\ref{sec:beautification} which optimizer yields a better solution.

\begin{algorithm}[t]
  \SetAlgoNoLine
  \KwIn{
    pre-trained generator~\del{$\bar{G}$}\add{$\hat{G}$},
    pre-trained discriminator~\del{$\bar{D}$}\add{$\hat{D}$},
    labels~$L$, constraints~$C$,
    initial Lagrange multipliers~$\mathbf{\lambda}^0$,
    initial penalty parameter~$\mu_0$
  }
  \KwOut{bounding boxes~$B^\ast$}

  \vspace{1mm}
  $Z_0 \leftarrow Z \sim \mathcal{N}(\mathbf{0}, \mathbf{I})$ \\
  $k \leftarrow 0$ \\

  \Repeat{stopping criteria is fulfilled}{
    \tcp{Inner optimization (Eq.~\eqref{eqn:auglag})}
    $Z^\ast \leftarrow
     \operatornamewithlimits{argmin}_Z L_A(Z\del{, \mathbf{\lambda}^k};\add{\mathbf{\lambda}^k,} \mu_k\add{, \hat{G}, \hat{D}, L, C})$
     \ starting at $Z_k$ \\[1mm]

    Update the Lagrange multipliers by Eq.~\eqref{eqn:update_lag} to obtain $\mathbf{\lambda}^{k+1}$ \\
    Update the penalty parameter by Eq.~\eqref{eqn:update_mu} to obtain $\mu_{k+1}$ \\
    $Z_k \leftarrow Z^\ast$ \\
    $k \leftarrow k + 1$ \\[1mm]
  }

  \vspace{1mm}
  $B^\ast \leftarrow \del{\bar{G}}\add{\hat{G}}(Z^\ast, L)$ \\
  \Return{$B^\ast$}

  \caption{Constrained layout generation via latent code optimization}
  \label{algo:generate_layout}
\end{algorithm}

In practice, optimizing the output value of the discriminator directly may
yield an adversarial example, \ie, the discriminator considers it as the real,
but perceptually degraded.  To avoid this, we clamp the output value of the
discriminator based on a certain threshold.  Specifically, we use $f(Z_0)$ as
the threshold, and $f'(Z) = \max\big(f(Z) - f(Z_0), 0\big)$ instead of $f(Z)$
in Eq.~\eqref{eqn:auglag}.

%% file: experiment.tex
\section{Experiments}

We evaluate the proposed method on both unconstrained and constrained layout
generation tasks.  We first describe the datasets and evaluation metrics, and then
explain the experimental setup for each task.

\subsection{Dataset}
% tab:dataset
\TabDataset{}
We evaluate layout generation on different types of graphic designs.  We use
three publicly available datasets:  Rico~\cite{Deka2017,Liu2018} provides UI
designs collected from mobile apps,  PubLayNet~\cite{Zhong2019} compiles a
dataset of document images, and Magazine~\cite{Zheng2019} collects magazine
pages.  Following the previous studies~\cite{Li2019,Lee2020}, we exclude
elements whose labels are not in the 13 most frequent labels in the Rico dataset, and
exclude layouts with more than 10 elements in both the Rico and PubLayNet datasets.  For
the PubLayNet dataset, we use 95\% of the official training split for training, the rest
for validation, and the official validation split for testing.  For Rico and
Magazine, since there is no official data split, we use 85\% of the dataset for
training, 5\% for validation, and 10\% for testing.  We summarize the
statistics of the datasets in Table~\ref{tab:dataset}.

\subsection{Evaluation Metrics}
We use four metrics to measure the quality of the generated layouts:
Fr\'{e}chet Inception Distance~(FID)~\cite{Heusel2017}, Maximum Intersection
over Union~(IoU), Alignment, and Overlap.

% tab:compare_fid
\TabFID{}

\subsubsection{FID}
To compute FID, we need to define the representative features of layouts.  We
follow the approach of \citet{Lee2020}, and train a neural network to
classify between real layouts and noise added layouts, and use the intermediate
features of the network.  One difference from \cite{Lee2020} is that we
incorporate the auxiliary decoder in Sec~\ref{sec:auxiliary_decoder} learning such
that the trained network is aware of both alignment and positions.
% We experimentally found that
% their features are sensitive to the local alignment of the layout,
% but less sensitive to global positional trends.
% We found that adding reconstruction loss to train the network mitigates this issue.
In Table~\ref{tab:compare_fid}, we show a comparison of FIDs across networks
learned with different objectives; \emph{Class} is real/fake classification
only, \emph{Recon} is auxiliary reconstruction only, and \emph{Class+Recon} is
learned with both objectives. The combination of both objectives
improves the sensitivity to different layout arrangements.

\subsubsection{Maximum IoU}
Maximum IoU is defined between two collections of generated layouts and
references.  We first define IoU based similarity between two layouts
$B=\{\mathbf{b_i}\}_{i=1}^N$ and $B'=\{\mathbf{b'_i}\}_{i=1}^N$.  We consider
the optimal matching between $B$ and $B'$, then compute the average IoU of
bounding boxes.  Let $\pi \in \mathcal{S}_N$ be a one-by-one matching, and
$\mathcal{S}_N$ be a set of possible permutations for size $N$.  Note that we
only consider matches between two bounding boxes with the same label, \ie, $l_i
= l_{\pi(i)} ( 1 \le i \le N)$.  The similarity with respect to the optimal
matching is computed as
\begin{align}
g_\mathrm{IoU}(B, B', L) &=
    \max_{\pi \in \mathcal{S}_N}
    \frac{1}{N} \sum_{i = 1}^N \mathrm{IoU}(\mathbf{b}_i, \mathbf{b}'_{\pi(i)}),
\end{align}
\noindent where $\mathrm{IoU}(\cdot, \cdot)$ computes IoU between bounding boxes.
To evaluate the similarity between generated layouts $\mathcal{B}\add{=\{B_m\}_{m=1}^M}$ and
references $\mathcal{B}'\add{=\{B'_m\}_{m=1}^M}$, we compute the average similarity on the optimal
matching:
\begin{align}
    \mathrm{MaxIoU}(\mathcal{B}, \mathcal{B}', \mathcal{L}) =
    \max_{\pi \in \mathcal{S}_{\del{D}\add{M}}}
    \frac{1}{\del{D}\add{M}} \sum_{m = 1}^{\del{D}\add{M}} g_\mathrm{IoU}(B_m, B'_{\pi(m)}, L_m),
\end{align}
\noindent where we only consider matches between two layouts with an identical
label set, \ie, $L_m = L_{\pi(m)} (1 \le m \le \del{D}\add{M})$.
% We define the Maximum IoU as between two collections of sets of bounding boxes.
% We denote the size of a collection as $D$,
% and two collections as $\mathcal{B} = \{ B_j \}_{j=1}^D$
% and $\mathcal{B}' = \{ B'_j \}_{j=1}^D$.
% In this metric, we consider a optimal assignment such that
% the IoU of two bounding boxes belonging to different collections is maximized.
% Let $\mathcal{L} = \{\!\!\{ L_j \}\!\!\}_{j=1}^D$ be a collection of multisets of labels,
% the Maximum IoU is computed as follows:
% \begin{align}
% g_\mathrm{IoU}(B, B', L) &=
%     \max_{\pi \in \bigg\{
%         \substack{
%         \pi | l_i = l_{\pi(i)} \\
%         \hphantom{\pi |} 1 \le i \le N \\
%         \hphantom{\pi |} \pi \in \mathcal{S}_N}
%         \bigg\} }
%     \frac{1}{N} \sum_{i = 1}^N \mathrm{IoU}(\mathbf{b}_i, \mathbf{b}'_{\pi(i)}) \\
% \mathrm{MaxIoU}(\mathcal{B}, \mathcal{B}', \mathcal{L}) &=
%     \max_{\pi \in \bigg\{
%         \substack{
%         \pi | L_m = L_{\pi(m)} \\
%         \hphantom{\pi |} 1 \le m \le D \\
%         \hphantom{\pi |} \pi \in \mathcal{S}_D}
%         \bigg\} }
%     \frac{1}{D} \sum_{m = 1}^D g_\mathrm{IoU}(B_m, B'_{\pi(m)}, L_m)
% \end{align}
% where $\pi$ is a permutation and $\mathcal{S}_X$ is a set of possible permutations for size $X$.
We use the solver~\cite{Crouse2016} provided by SciPy~\cite{Virtanen2020} to
solve the assignment problems.

\subsubsection{Alignment and overlap}
We use the \emph{Alignment} and \emph{Overlap} metrics used in the previous
work~\cite{Li2020}.  We modify the original metrics by normalizing with the
number of elements $N$.

\subsection{Unconstrained Layout Generation}
\label{sec:ulg_experiment}

\subsubsection{\del{Baselines}\add{Setup}}
We use LayoutGAN~\cite{Li2019} and NDN~\cite{Lee2020} as baselines.  Although
LayoutGAN is intended for the unconditional setting, we adapt the model to
be conditioned on a label set input.  We refer to the model using the wireframe
rendering discriminator as {\bf LayoutGAN-W} and the one using the
relation-based discriminator as {\bf LayoutGAN-R}.  NDN first generates the
position and size relations between elements, then generates bounding boxes
based on the relations, and finally modifies the misalignment of the boxes.  We
denote it as {\bf NDN-none} to match the designation in their paper, as our
setting does not specify the relations.  We reimplement all the baselines as
since the official codes for the baselines are not publicly
available\footnote{ The authors of LayoutGAN provide only the code for point
layout experiment in
\url{https://github.com/JiananLi2016/LayoutGAN-Tensorflow}, not for bounding
boxes.  }.  We implement our LayoutGAN++ with PyTorch~\cite{pytorch_paper}.  We
train the model using the Adam optimizer with 200,000 iterations with a batch
size of 64 and a learning rate of 1e-5, taking \del{half a day}\add{six hours}
with a GPU of NVIDIA GeForce RTX 2080Ti.
\add{Our Transformer modules consist of 8 blocks, and in}\del{For}
each \del{Transformer} block, we set the input/output dimension to 256, the dimension
of the hidden layer to 128, and the number of multi-head attentions to
\del{8}\add{4}.

\subsubsection{Results}

% tab:ulg_result
\ULGTable{}

% fig:ulg_result
\ULGFigure{}

We summarize the quantitative comparison in Table~\ref{tab:ulg_result} and the
qualitative comparison in Fig.~\ref{fig:ulg_result}.  Since all the comparison
methods are stochastic, we report the mean and standard deviation of five
evaluations with the same trained model.  Regarding LayoutGAN~\cite{Li2019}, we
find that LayoutGAN-W is unstable to train, and failed to reproduce the results
as good as in their paper despite our efforts, which is similarly reported in
the recent studies~\cite{Gupta2021,Arroyo2021}.  Our results show that
LayoutGAN-R is much stable to train, and outperforms LayoutGAN-W.  Our
LayoutGAN++ achieves comparable to or better results than the current
state-of-the-art method NDN-none~\cite{Lee2020}, in particular, results on the
Rico dataset are similar, while results on the PubLayNet dataset and Magazine
dataset are favourable to our approach.

% \ULGAblTable{}

\subsection{Layout Generation with Beautification Constraints} \label{sec:beautification}
The goal of this setting is to generate a well-aligned layout with no
overlapping, which can serve as a post-processing to beautify the result of the
unconstrained layout generation.  We conduct the experiment with the PubLayNet
dataset, in which most of the layouts are aligned and have little overlap.

\subsubsection{Constraints}
Let $g_\mathrm{align}$ be the function that computes the Alignment metric,
we express the alignment constraint as
\begin{equation}
    c_\mathrm{align}(B) = \max \big( g_\mathrm{align}(B)\del{, \tau}\add{-\tau, 0} \big) \ ,
\end{equation}
\noindent where $\tau$ is a threshold parameter. We set $\tau=0.004$ in our
experiment.  We use the Overlap metric as the non-overlapping constraint
$c_\mathrm{ovrlp}$.

\subsubsection{\del{Baselines}\add{Setup}}
We use a pre-trained LayoutGAN++ model within our proposed CLG-LO framework to
perform the constrained task.  We follow the same settings as in Section
\ref{sec:ulg_experiment} for training LayoutGAN++.  We compare two different
inner optimizers, Adam~\cite{Kingma2015} and CMA-ES~\cite{Hansen2016}.
\add{The mean runtime for CLG-LO was 13.6 seconds with Adam (SD: 11.2) and 1.45
seconds with CMA-ES (SD: 1.75).}

Since there is no directly comparable methods in the literature for this
setting, we design a baseline called {\bf CAL} that uses constraints as
additional losses, referring to the similar work~\cite{Li2020}.  To instantiate
CAL, we train LayoutGAN++ with both the alignment constraint $c_\mathrm{align}$
and the non-overlapping constraint $c_\mathrm{ovrlp}$ added to the generator
objective, which encourages a generated layout that satisfies the constraints,
but does\del{n't} \add{not} explicitly enforce them.

\subsubsection{Results}
% tab:clg_beautify
\CLGBTable{}

% fig:clg_beautify
\CLGBFigure{}

We summarize the quantitative comparison in Table~\ref{tab:clg_beautify}.  The base
model is LayoutGAN++ without beautification.  We can see that CAL performs better
in terms of Alignment and Overlap than the baseline, thanks to the added
losses.  FID and Maximum IoU are also improved, which may be due to the
inductive bias expressed as the added losses, making GAN easier to train.  Our
CLG-LO further improves Alignment and Overlap significantly with almost no
degradation in terms of FID and Maximum IoU.  As for the choice of inner
optimizer, CMA-ES seems to perform better than Adam.  We suspect that due to the
augmented Lagrangian function (Eq.~\eqref{eqn:auglag}) having many local
solutions, and thus a population-based global gradient-free optimization
method, \eg, CMA-ES, is more suitable than a gradient-based method, \eg, Adam.

We show the optimization results by CLG-LO using CMA-ES as the inner optimizer
in Fig.~\ref{fig:clg_beautify}. Our framework successfully found aligned and
non-overlapping layouts.  We have set the initial sigma parameter of CMA-ES
smaller to explore around the initial latent code, which leads to the optimized
layout not changing significantly from the initial layout.
% otani: This explanation about errors is not clear for me.
% We can also observe that some of the elements without violating the constraints are moved.
% Additional constraints may be needed to prevent it for more strict cases.

\subsection{Layout Generation with Relational Constraints}
In this setting, we consider a scenario where the user specifies the location
and size relationships of elements in the layout.  We consider three size
relations, \textit{smaller}, \textit{larger} and \textit{equal}, and five
location relations, \textit{above}, \textit{bottom}, \textit{left},
\textit{right}, and \textit{overlap}.  We also define the relation to the
canvas, \eg, positioning at the top of the canvas.  We determine the relations
from the ground-truth layout and use its subset as constraines. We change
percentages of the relations used as constraints and report the rate of
violated constraints.

% fig:clg_relation
\CLGRFigure{}

% tab:clg_relation
\CLGRTable{}

% fig:clg_plot
\CLGRPlot{}

\subsubsection{Constraints}
The size constraint~$c_\mathrm{size}$ is defined as the sum of cost functions
of all size relations.  For example, suppose the user specifies that the $j$-th
element has to be larger than the $i$-th element, then the cost function of
\textit{larger} relation is defined by:
\begin{equation}
    g_\mathrm{lg}(\mathbf{b}_i, \mathbf{b}_j) = \max \big( (1 + \del{r}\add{\gamma}) a(\mathbf{b}_i) - a(\mathbf{b}_j), 0 \big),
\end{equation}
\noindent where $a(\cdot)$ is a function that calculates the area of a given bounding
box, and $\del{r}\add{\gamma}$ is a tolerance parameter shared across the size relations. We set
$r=0.1$ in our experiment.

We also define the location constraint~$c_\mathrm{loc}$ in the same way.  For
example, suppose the user specifies that the $j$-th element has to be above the
$i$-th element, then the cost function of \textit{above} relation is defined
by:
\begin{equation}
    g_\mathrm{ab}(\mathbf{b}_i, \mathbf{b}_j) = \max \big( y_\mathrm{b}(\mathbf{b}_j) - y_\mathrm{t}(\mathbf{b}_i), 0 \big),
\end{equation}
\noindent where $y_\mathrm{t}(\cdot)$ and $y_\mathrm{b}(\cdot)$ are functions that return the
top and bottom coordinates of a given bounding box, respectively.

\subsubsection{\del{Baselines}\add{Setup}}
We compare our CLG-LO against NDN~\cite{Lee2020}.  In CLG-LO, we use CMA-ES for
the inner optimizer, as it worked well in the experiments with beautification
constraints.  The rest of the settings follow the experiment with
beautification constraints\add{, but for a fair comparison, we did not use the
beautification constraints themselves. The mean runtime for CLG-LO was 1.96
seconds (SD: 3.48).}

\subsubsection{Results}

We show the qualitative results in Fig.~\ref{fig:clg_relation} and the
quantitative comparison in Table~\ref{tab:clg_relation}.
We report the results for a setting that uses
10\% of all relations in Table~\ref{tab:clg_relation}, which is what we believe would be representative of a
realistic usage scenario. A typical example that uses roughly 10\% relations is
the upper left one in Fig.~\ref{fig:clg_relation}.  Our CLG-LO performed
comparable to or better than NDN, and in particular showed significant
improvement in  the constraint violation metric.  This is as to be expected
because NDN does not guarantee the inferred result satisfies the constraints,
whereas our method tries to find a solution that satisfies as many of the
constraints as possible through iterative optimization.

We also show in Fig.~\ref{fig:clg_plot} the experimental results of varying the
percentage of relations used.  We can find that NDN performs better as
increasing the number of relations used, which is reasonable since its layout
generation module is trained with the complete relational graph of the
ground-truth layout.  On the other hand, our CLG-LO performs unfavorably as
increasing the number of relations used, because it becomes harder to find a
solution that satisfies the constraints.  A practical remedy when no solution
is found could be to store a layout for each iteration of the main loop in
Algorithm~\ref{algo:generate_layout}, and let the user choose one based on the
trade-off between constraint satisfaction and layout quality.  We note,
however, that our method performs best in realistic scenarios where the number
of user-specified relations is few.

%% file: conclusion.tex
\section{Conclusions and Discussion}

In this paper, we proposed a novel framework called Constrained Layout
Generation via Latent Optimization (CLG-LO), which performs constrained layout
generation by optimizing the latent codes of pre-trained GAN.  While existing
works treat constraints as either additional objectives or conditioning,
requiring re-training when unexpected constraints are involved, our framework can
flexibly incorporate a variety of constraints using a single unconstrained GAN.
While our approach is applicable to most generative layout design models, we
also present a new layout generation model called LayoutGAN++ that is able to
outperform existing approaches in unconditioned generation.  Experimental
results on both unconstrained and constrained generation tasks using three
public datasets support the effectiveness of the proposed methods.

While our approach is able to significantly outperform existing approaches in
many cases\del{. G}\add{, g}iven the non-convexity and complexity of the optimization problem
as the objective \add{and constraint} function\add{s} in Eq.~\eqref{eqn:const}
\del{is}\add{involve} a complex nonlinear neural network, we have no guarantees
on the convergence of the approach. When
the number of constraints becomes large (Figure~\ref{fig:clg_plot}), the
optimizer can have issues finding a good solution, and underperform existing
approaches. However, in general, most users will not specify very large number\del{s}
of constraints, and in those situations, our approach significantly outperforms
existing approaches. We believe that this effect can be mitigated by improving
the optimization approach itself, using piece-wise convex approximations, or
improving the initialization of the optimization variables.
\add{It may also be practical to design an interaction that asks the user to
remove or change difficult constraints.}

\add{
Our optimization-based approach allows us to flexibly change not only the
constraint function, but also the objective function.  For example, if we wish to
limit the amount of change, we can add the distance between the boxes before
and after the optimization as a penalty to the objective function.
Our approach can also be applied to any model that can generate diverse
plausible layouts through manipulating latent variables.
Note that when used with VAE-based models~\cite{Jyothi2019,Arroyo2021,Lee2020}
that do not have an explicit function
to measure the quality of the generated layout, it becomes a constraint
satisfaction problem. Our approach still works in such cases, but if the quality
of the outcome is problematic, it may be necessary to train an additional
measurement network like a discriminator.
}

There are many open directions for improvement such as incorporating models
that approximate human perception as constraints~\cite{Bylinskii2017,Zhao2018}
in order to generate more aesthetically pleasing results.  Exploring latent
codes considering the diversity of layouts is another exciting
direction~\cite{Pugh2016}, allowing for efficient design exploration with a variety of alternatives.
Also, it is worth investigating whether or not our proposed CLG-LO approach can be
applied generation problems other than that of layout designs.

\begin{acks}
    % The authors would like to thank the anonymous referees for their service.
    \add{This work is partially supported by Waseda University Leading Graduate
    Program for Embodiment Informatics.}
\end{acks}

%% file: ms.bbl
%%% -*-BibTeX-*-
%%% Do NOT edit. File created by BibTeX with style
%%% ACM-Reference-Format-Journals [18-Jan-2012].